\documentclass[10pt,twocolumn,letterpaper]{article}

\usepackage{cvpr}
\usepackage{times}
\usepackage{epsfig}
\usepackage{graphicx}
\usepackage{amsmath}
\usepackage{amssymb}
\usepackage{comment}
\usepackage[font=small]{caption} %
\usepackage{booktabs} %
\usepackage{multirow} %

\usepackage{cuted}
\usepackage{capt-of}

\usepackage{siunitx}

\newcommand{\mypar}[1]{\vspace{-2mm}\paragraph{#1}}
\newcommand{\fig}[1]{Figure~\ref{#1}}

\newcommand{\sect}[1]{Section~\ref{#1}}

\newcommand{\unet}{UNet} %

\usepackage[usenames]{color}
\usepackage[usenames,dvipsnames]{xcolor}

\newcommand {\note}[1]{}
\newcommand {\todo}[1]{}
\newcommand {\malik}[1]{}
\newcommand {\shiry}[1]{}
\newcommand {\amir}[1]{}
\newcommand {\gefen}[1]{}
\newcommand {\caroline}[1]{}
\newcommand {\owens}[1]{}

\usepackage[pagebackref=true,breaklinks=true,letterpaper=true,colorlinks,bookmarks=false]{hyperref}

\cvprfinalcopy %

\def\cvprPaperID{1766} %

\ifcvprfinal\fi
\begin{document}

\title{Learning Individual Styles of Conversational Gesture}

\author{Shiry Ginosar\thanks{Indicates equal contribution.}\\
UC Berkeley\\
\and
Amir Bar\footnotemark[1] \\
Zebra Medical Vision\\
\and
Gefen Kohavi\\
UC Berkeley\\
\and
Caroline Chan\\
MIT \\
\and
\and
Andrew Owens\\
UC Berkeley\\
\and
Jitendra Malik\\
UC Berkeley\\
}

\maketitle

\begin{strip}
\centering
\vspace{-6mm}
\includegraphics[width=\textwidth]{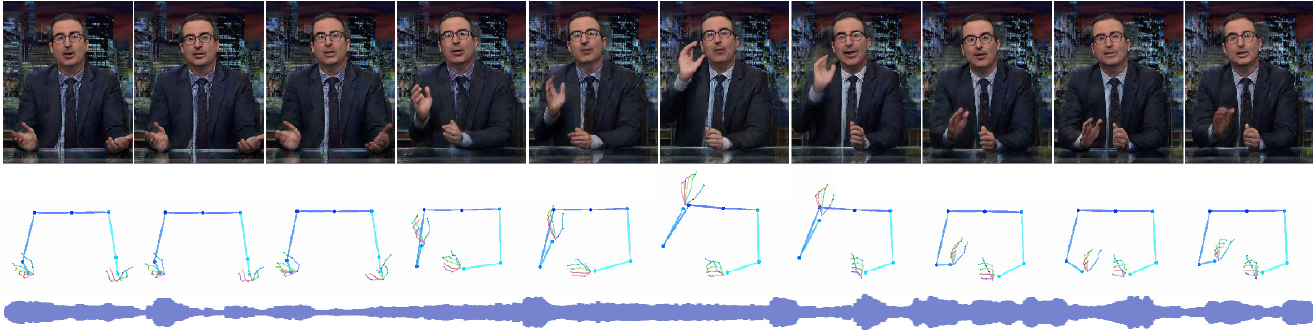}
\vspace{-6mm}
\captionof{figure}{{\bf Speech-to-gesture translation example}. In this paper, we study the connection between conversational gesture and speech. Here, we show the result of our model that predicts gesture from audio. From the bottom upward: the input audio, arm and hand pose predicted by our model, and video frames synthesized from pose predictions using~\cite{Chan2018dance}. 
(See \small{\url{http://people.eecs.berkeley.edu/~shiry/speech2gesture}} for video results.)}
\label{fig:teaser}
\end{strip}

\begin{abstract}
Human speech is often accompanied by hand and arm gestures.
Given audio speech input, we generate plausible gestures to go along with the sound.
Specifically, we perform cross-modal translation from ``in-the-wild'' monologue speech of a
single speaker to their hand and arm motion. 
We train on unlabeled videos for which we only have noisy pseudo ground truth from an automatic pose detection system.
Our proposed model significantly outperforms baseline methods in a quantitative comparison.
To support research toward obtaining a computational understanding of the relationship between gesture and speech, we release a large video dataset of person-specific gestures.
\end{abstract}

\section{Introduction}

\begin{figure*}
\centering
  \includegraphics[width=1.0\linewidth]{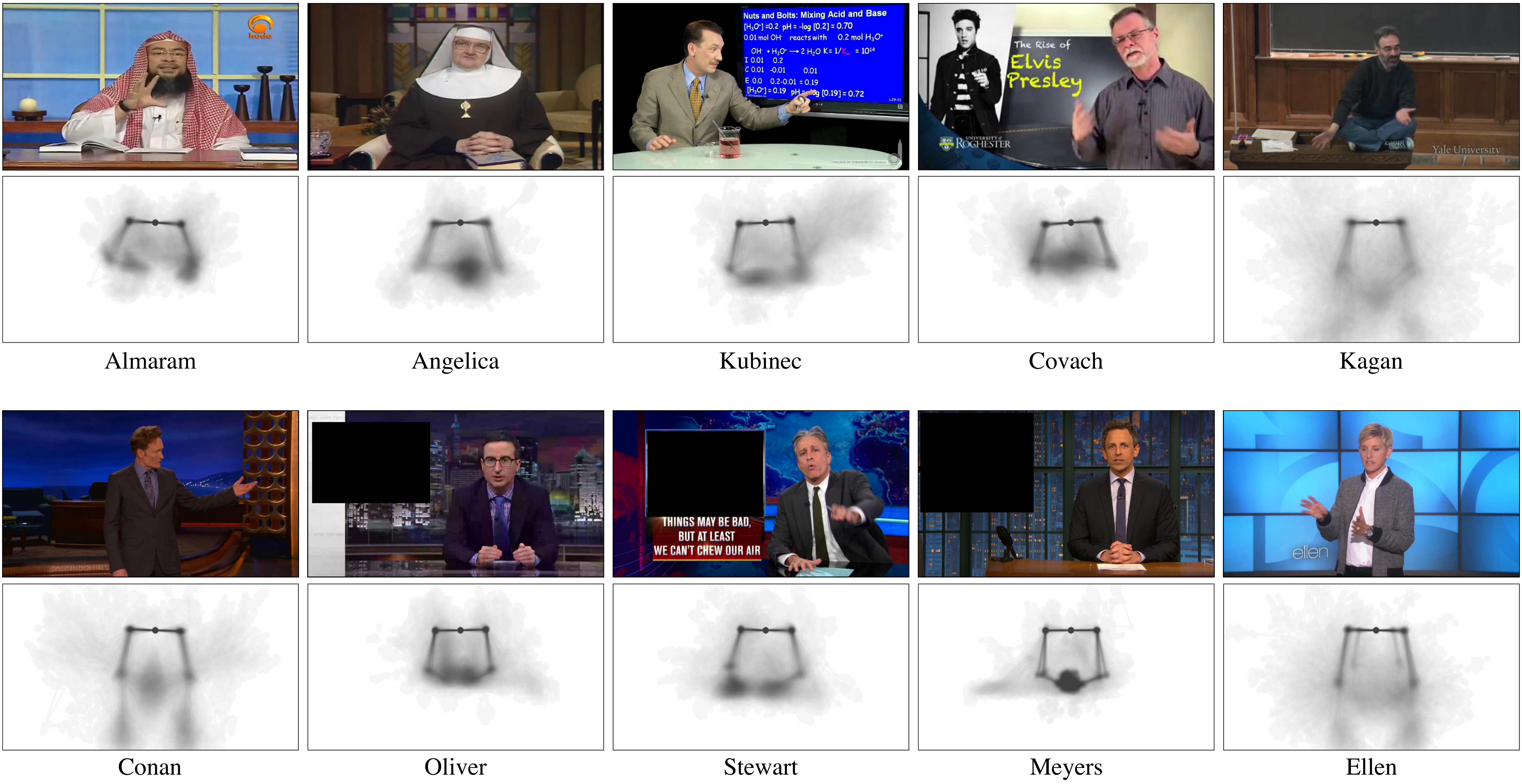}
\vspace{-4mm}
  \caption{{\em Speaker-specific gesture dataset}. We show a representative video frame for each speaker in our dataset. Below each one is a heatmap depicting the frequency that their
  arms and hands appear in different spatial locations (using the skeletal representation of gestures shown in \fig{fig:teaser}). This visualization reveals the speaker's resting pose, and how they tend to move---for example, {\em Angelica} tends to keep her hands folded, whereas {\em Kubinec} frequently points towards the screen with his left hand. Note that some speakers, like {\em Kagan}, {\em Conan} and {\em Ellen}, alternate between sitting and standing and thus the distribution of their arm positions is bimodal.}
\label{fig:data-stats}
\end{figure*}

When we talk, we convey ideas via two parallel channels of communication---speech and
gesture. These conversational, or co-speech, gestures are the hand and arm motions we
spontaneously emit when we speak~\cite{McNeill92}. They complement speech and add non-verbal information that help our listeners comprehend what we say~\cite{Cassell-listeners}.
Kendon~\cite{kendon_2004} places conversational gestures at one end of a continuum, with sign language, a true language, at the other end. In between the
two extremes are pantomime and emblems like ``Italianite'', with an agreed-upon vocabulary and culture-specific meanings. A gesture can be subdivided into phases describing its progression from the speaker's rest position, through the gesture preparation, stroke, hold and retraction back to rest.

Is the information conveyed in speech and gesture correlated? This is a topic of ongoing debate. The {\em hand-in-hand} hypothesis claims that gesture is redundant to speech when speakers refer to subjects and objects in scenes~\cite{hand-in-hand}. In contrast, according to the {\em trade-off hypothesis}, speech and gesture are complementary since people use gesture when speaking would require more effort and vice versa~\cite{tradeoff}. We approach the question from a data-driven learning perspective and ask to what extent can we predict gesture motion from the raw audio signal of speech.

We present a method for \emph{temporal cross-modal translation}. Given an input audio clip of a spoken statement (Figure~\ref{fig:teaser} bottom), we generate a corresponding motion of the speaker's arms and hands which matches the style of the speaker, despite the fact that we have never seen or heard this person say this utterance in training (Figure~\ref{fig:teaser} middle). We then use an existing video synthesis method to visualize what the speaker might have looked like when saying these words (Figure~\ref{fig:teaser} top).

To generate motion from speech, we must learn a mapping between audio and pose. While this can be formulated as translation, in practice there are two inherent challenges to using the natural pairing of audio-visual data in this setting. First, gesture and speech are 
\emph{asynchronous}, as gesture can appear before, after or during the corresponding utterance~\cite{Butterworth}. 
Second, this is a \emph{multimodal} prediction task as speakers may perform different gestures while saying the same thing on different occasions. Moreover, acquiring human annotations for large amounts of video is infeasible. We therefore need to get a training signal from \emph{pseudo ground truth} of $2D$ human pose detections on unlabeled video.

Nevertheless, we are able to translate speech to gesture in an end-to-end fashion from the
raw audio to a sequence of poses. To overcome the asynchronicity issue we use a large temporal context (both past and future) for prediction. Temporal context also allows for smooth gesture prediction despite the noisy automatically-annotated pseudo ground truth. Due to multimodality, we do not expect our predicted motion to be the same as the ground truth. However, as this is the only training signal we have, we still use automatic pose detections for learning through regression. To avoid regressing to the mean of all modes, we apply an adversarial discriminator~\cite{gan} to our predicted motion. This ensures that we produce motion that is ``real" with respect to the current speaker.

Gesture is idiosyncratic~\cite{McNeill92}, as different speakers tend to use different styles
of motion (see Figure~\ref{fig:data-stats}). It is therefore important to learn a personalized gesture model for each speaker. To address this, we present a large, $144$-hour \textit{person-specific} video dataset of $10$ speakers that we make publicly available\footnote{\url{http://people.eecs.berkeley.edu/~shiry/speech2gesture}}. We deliberately pick a set of speakers for which we can find hours of clean single-speaker footage. Our speakers come from a diverse set of backgrounds: television show hosts, university lecturers and televangelists. They span at least three religions and discuss a large range of topics from commentary on current affairs through the philosophy of death, chemistry and the history of rock music, to readings in the Bible and the Qur'an.

\section{Related Work}
\label{sec:related}
\paragraph{Conversational Gestures}

McNeill~\cite{McNeill92} divides gestures into several classes~\cite{McNeill92}: \textit{emblematics} have specific conventional meanings (\eg ``thumbs up!''); \textit{iconics} convey physical shapes or direction of movements; \textit{metaphorics} describe abstract content using concrete motion; \textit{deictics} are pointing gestures, and \textit{beats} are repetitive, fast hand motions that provide a temporal framing to speech. 

Many psychologists have studied questions related to co-speech gestures~\cite{McNeill92,kendon_2004} (See~\cite{WAGNER2014} for a review). This vast body of research has mostly relied on studying a small number of individual subjects using recorded choreographed story retelling in lab settings. Analysis in these studies was a manual process. Our goal, instead, is to study conversational gestures in the wild using a data-driven approach.

Conditioning gesture prediction on speech is arguably an ambiguous task, since gesture and speech may not be synchronous. While McNeill~\cite{McNeill92} suggests that gesture and speech originate from a common source and thus should co-occur in time according to well-defined rules, Kendon~\cite{kendon_2004} suggests that gesture starts before the corresponding utterance. Others even argue that the temporal relationships between speech and gesture are not yet clear and that gesture can appear before, after or during an utterance~\cite{Butterworth}.

\mypar{Sign language and emblematic gesture recognition} There has
been a great deal of computer vision work geared towards recognizing
sign language gestures from video. This includes methods that use video transcripts as a weak source of
supervision~\cite{buehler2009learning}, as well as recent methods based on
CNNs~\cite{pfister2014deep,koller2016deep} and RNNs~\cite{camgoz2018sign}.  There has also been work that
recognizes emblematic hand and face gestures
\cite{freeman1995orientation,Darrell-Irfan-Essa}, head gestures \cite{morency2007latent}, and co-speech gestures~\cite{quek2002multimodal}. By contrast, our goal is to predict co-speech gestures from audio.

\mypar{Conversational agents} 
Researchers have proposed a number of methods for generating plausible
gestures, particularly for applications with conversational
agents~\cite{cassell2000embodied}. In early work, Cassell
\etal~\cite{Cassell-animated} proposed a system that guided arm/hand
motions based on manually defined rules. Subsequent rule-based
systems~\cite{kopp2006towards} proposed new ways of expressing
gestures via annotations. 

More closely related to our approach are
methods that learn gestures from speech and text, without requiring an
author to hand-specify rules. Notably,~\cite{cassell2004beat}
synthesized gestures using natural language processing of spoken text,
and Neff~\cite{Kipp-modelling-synthesis} proposed a system for making person-specific gestures. Levine
\etal~\cite{levine2009real} learned to map acoustic prosody features to motion using a HMM. Later
work~\cite{levine2010gesture} extended this approach to use
reinforcement learning and speech recognition, combined acoustic analysis with text~\cite{marsella2013performance}, created hybrid rule-based systems~\cite{Busso-meaningful}, and used restricted Boltzmann machines for inference~\cite{chiu2011train}. 
Since the goal of these methods is to generate motions for virtual agents, they use lab-recorded audio, text, and motion
capture. This allows them to use simplifying assumptions that present challenges for in-the-wild video analysis like ours: \eg,~\cite{levine2009real} requires precise 3D pose and assumes
that motions occur on syllable boundaries, and~\cite{chiu2011train} assumes that gestures are initiated by an upward motion of the wrist. In contrast with these methods, our approach does not explicitly use any
text or language information during training---it learns gestures from
raw audio-visual correspondences---nor does it use hand-defined
gesture categories: arm/hand pose are predicted directly from audio.

\mypar{Visualizing predicted gestures}
One of the most common ways of visualizing gestures
is to use them to animate a 3D avatar~\cite{thiebaux2008smartbody,levine2010gesture,hartholt2013all}. Since our work studies personalized gestures for in-the-wild videos, where 3D data is not available, we use a data-driven synthesis approach inspired by Bregler \etal~\cite{bregler1997video}. To do this, we employ the pose-to-video method of Chan \etal~\cite{Chan2018dance}, which uses a conditional generative adversarial network (GAN) to synthesize videos of human bodies from pose.

\mypar{Sound and vision}
Aytar \etal~\cite{aytar2016soundnet} use the synchronization of visual and audio signals in natural phenomena to  learn sound representations from unlabeled in-the-wild videos. To do this, they transfer knowledge from trained discriminative models in the visual domain, to the audio domain. 

Synchronization of audio and visual features can also be used for synthesis. Langlois \etal~\cite{Langlois2014} try to optimize for synchronous events by generating rigid-body animations of objects falling or tumbling that temporally match an input sound wave of the desired sequence of contact events with the ground plane. More recently, Shlizerman \etal~\cite{shlizermanaudio} animated the hands of a 3D avatar according to input music. However, their focus was on music performance, rather than gestures, and consequently the space of possible motions was limited (\eg, the zig-zag motion of a violin bow). Moreover, while music is uniquely defined by the motion that generates it (and is synchronous with it), gestures are neither unique to, nor synchronous with speech utterances.

Several works have focused on the specific task of synthesizing videos of faces speaking, given audio input.
Chung \etal~\cite{Chung17b} generate an image of a talking face from a still image of the speaker and an input speech segment by learning a joint embedding of the face and audio. Similarly, ~\cite{Obama2017} synthesizes videos of Obama saying novel words by using a recurrent neural network to map speech audio to mouth shapes and then embedding the synthesized lips in ground truth facial video. While both methods enable the creation of fake content by generating faces saying words taken from a different person, we focus on single-person models that are optimized for animating same-speaker utterances. Most
importantly, generating gesture, rather than lip motion, from speech is more involved as gestures are asynchronous with speech, multimodal and person-specific.

\section{A Speaker-Specific Gesture Dataset}

\label{sec:data}
\newcommand{\dsetname}{Gestures\xspace}

We introduce a large $144$-hour video dataset specifically tailored to studying speech and gesture of individual speakers in a data-driven fashion. As shown in \fig{fig:data-stats}, our dataset contains in-the-wild videos of $10$ gesturing speakers that were originally recorded for television shows or university lectures. We collect several hours of video per speaker, so that we can individually model each one. We chose speakers that cover a wide range of topics and gesturing styles. Our dataset contains: $5$ talk show hosts, $3$ lecturers and $2$ televangelists. Details about data collection and processing as well as an analysis of the individual styles of gestures can be found in the supplementary material.

\mypar{Gesture representation and annotation}  We represent the speakers' pose over time using a temporal stack of 2D skeletal keypoints, which we obtain using OpenPose~\cite{cao2017realtime}. From the complete set of keypoints detected by OpenPose, we use the $49$ points corresponding to the neck, shoulders, elbows, wrists and hands to represent gestures. Together with the video footage, we provide the skeletal keypoints for each frame of the data at a $15$fps. Note, however, that these are not ground truth annotations, but a proxy for the ground truth from a state-of-the-art pose detection system.

\mypar{Quality of dataset annotations} All ground truth, whether from human observers or otherwise, has associated error. The pseudo ground truth we collect using automatic pose detection may have much larger error than human annotations, but it enables us to train on much larger amounts of data. Still, we must estimate whether the accuracy of the pseudo ground truth is good enough to support our quantitative conclusions. We compare the automatic pose detections to labels obtained from human observers on a subset of our training data and find that the pseudo ground truth is close to human labels and that the error in the pseudo ground truth is small enough for our task. The full experiment is detailed in our supplementary material.

\section{Method}
\label{sec:method}
Given raw audio of speech, our goal is to generate the speaker's corresponding arm and hand gesture motion. We approach this task in two stages---first, since the only signal we have for training are corresponding audio and pose detection sequences, we learn a mapping from speech to gesture using $L_1$ regression to temporal stacks of $2D$ keypoints. Second, to avoid regressing to the mean of all possible modes of gesture, we employ an adversarial discriminator that ensures that the motion we produce is plausible with respect to the typical motion of the speaker.

\subsection{Speech-to-Gesture Translation}
\label{sec:translation-arch}

\begin{figure}
\centering
\includegraphics[width=\linewidth]{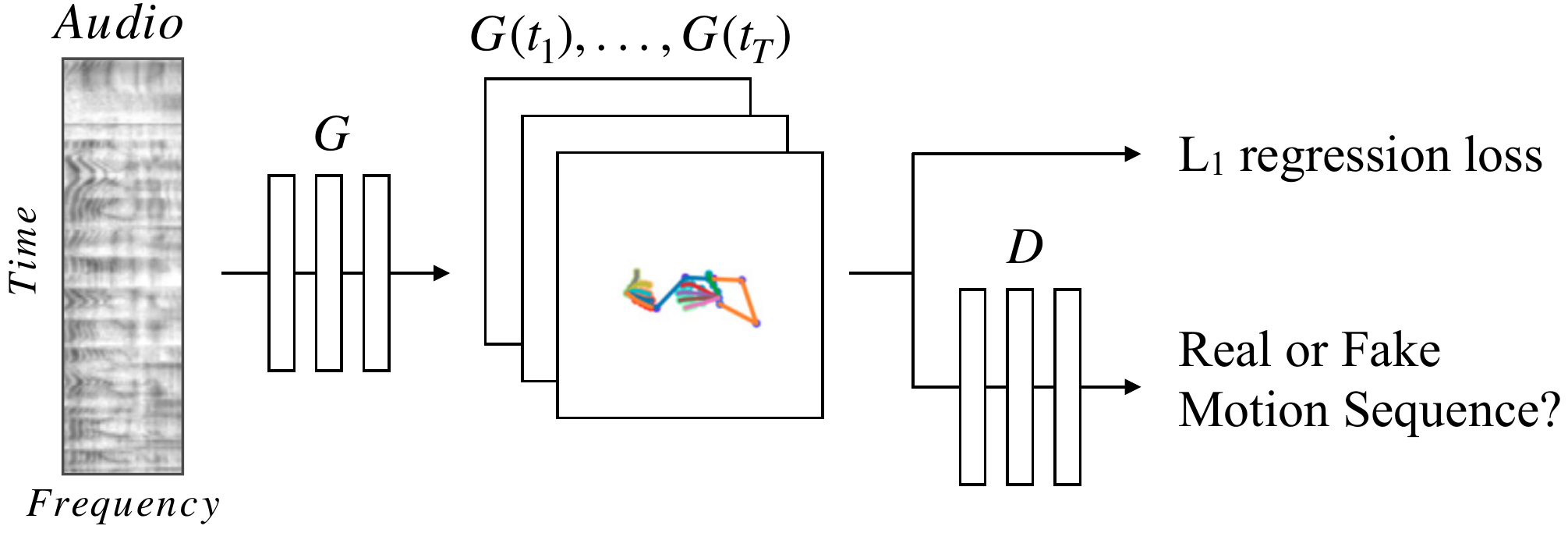}
\vspace{-4mm}
  \caption{\emph{Speech to gesture translation model.} A convolutional audio encoder downsamples the $2D$ spectrogram and transforms it to a $1D$ signal. The translation model, $G$, then predicts a corresponding temporal stack of $2D$ poses. $L_1$ regression to the ground truth poses provides a training signal, while an adversarial discriminator, $D$, ensures that the predicted motion is both temporally coherent and in the style of the speaker.}
\label{fig:translation-arch}
\end{figure}

Any realistic gesture motion must be temporally coherent and smooth. We accomplish smoothness by learning an audio encoding which is a representation of the whole utterance, taking into account the full temporal extent of the input speech, $\mathbf{s}$, and predicting the whole temporal sequence of corresponding poses, $\mathbf{p}$, at once (rather than recurrently).

Our fully convolutional network consists of an audio encoder followed by a $1D$ \unet{}~\cite{unet,pix2pix2017} translation architecture, as shown in
Figure~\ref{fig:translation-arch}. The audio encoder takes a $2D$ log-mel spectrogram as input, and downsamples it through a series of convolutions, resulting in a $1D$ signal with the same sampling rate as our video ($15$ Hz). The \unet{} translation architecture then learns to map this signal to a temporal stack of pose vectors (see \sect{sec:data} for details of our gesture representation) via an $L_1$ regression loss:
\begin{equation}
\mathcal{L}_{L_1}(G) = \mathbb{E}_{\mathbf{s},\mathbf{p}}[||\mathbf{p}-G(\mathbf{s})||_1].
\end{equation}
We use a \unet{} architecture for translation since its bottleneck provides the network with past and future temporal context, while the skip connections allow for high frequency temporal information to flow through, enabling prediction of fast motion.

\subsection{Predicting Plausible Motion}
While $L_1$ regression to keypoints is the only way we can extract a training signal from our data, it suffers from the known issue of regression to the mean which produces overly smooth motion. This can be seen in our supplementary video results. To combat the issue and ensure that we produce realistic motion, we add an adversarial discriminator~\cite{pix2pix2017,Chan2018dance} $D$, conditioned on the difference of the predicted sequence of poses. i.e.
the input to the discriminator is the vector $\mathbf{m}=[p_2-p_1, \dots , p_T-p_{T-1}]$ where $p_i$ are $2D$ pose keypoints and $T$ is the temporal extent of the input audio and predicted pose sequence. The discriminator $D$ tries to maximize the following objective while the generator $G$ (translation architecture, Section~\ref{sec:translation-arch}) tries to minimize it:
\begin{equation}
\mathcal{L}_{GAN}(G, D) = \mathbb{E}_\mathbf{m}[\log D(\mathbf{m})]+\mathbb{E}_\mathbf{s}[\log(1-G(\mathbf{s}))],
\end{equation}
where $s$ is the input audio speech segment and $m$ is the motion derivative of the predicted stack of poses. Thus, the generator learns to produce real-seeming speaker motion while the discriminator learns to classify whether a given motion sequence is real. Our full objective is therefore:
\begin{equation}
\min_G \max_D \mathcal{L}_{GAN}(G,D) + \lambda \mathcal{L}_{L_1}(G).
\end{equation}

\subsection{Implementation Details}
We obtain translation invariance by subtracting (per frame) the neck keypoint location from all other keypoints in our pseudo ground truth gesture representation (section ~\ref{sec:data}). We then normalize each keypoint (\eg left wrist) across all frames by subtracting the per-speaker mean and dividing by the standard deviation. During training, we take as input spectrograms corresponding to about $4$ seconds of audio and predict $64$ pose vectors, which correspond to about $4$ seconds at a $15$Hz frame-rate. At test time we can run our network on arbitrary audio durations. We optimize using Adam~\cite{adam} with a batch size of $32$ and a learning rate of \num{e-4}.
We train for 300K/90K iterations with and without an adversarial loss, respectively, and select the best performing model on the validation set.

\section{Experiments}
We show that our method produces motion that quantitatively outperforms several baselines, as well as a previous method that we adapt to the problem.

\subsection{Setup}
We describe our experimental setup including our baselines for comparison and evaluation metric.

\vspace{-3mm}
\subsubsection{Baselines}
\label{sec:baselines}
We compare our method to several other models. 
\vspace{-3mm}
\mypar{Always predict the median pose} Speakers spend most of their time in rest position~\cite{kendon_2004}, so predicting the speaker's median pose can be a high-quality baseline. For a visualization of each speaker's rest position, see Figure~\ref{fig:data-stats}.
\vspace{-3mm}
\mypar{Predict a randomly chosen gesture} In this baseline, we randomly select a different gesture sequence (which does not correspond to the input utterance) from the training set of the same speaker, and use this as our prediction. While we would not expect this method to perform well quantitatively, there is reason to think it would generate qualitatively appealing motion: these are real speaker gestures---the only way to tell they are fake is to evaluate how well they corresponds to the audio.
\vspace{-3mm}
\mypar{Nearest neighbors} Instead of selecting a completely random gesture sequence from the same speaker, we can use audio as a similarity cue. For an input audio track, we find its nearest neighbor for the speaker using pretrained audio features, and transfer its corresponding motion. To represent the audio, we use the state-of-the-art VGGish feature embedding~\cite{vggish} pretrained on AudioSet~\cite{gemmeke2017audio}, and use cosine distance on normalized features. 
\vspace{-3mm}
\mypar{RNN-based model~\cite{shlizermanaudio}} We further compare our motion prediction to an RNN architecture proposed by Shlizerman \etal. Similar to us, Shlizerman \etal predict arm and hand motion from audio in a $2D$ skeletal keypoint space. However, while our model is a convolutional neural network with log-mel spectrogram input, theirs uses a 
$1$-layer LSTM model that takes MFCC features (a low-dimensional, hand-crafted audio feature representation) as input.
We evaluated both feature types and found that for~\cite{shlizermanaudio}, MFCC features outperform the log-mel spectrogram features on all speakers. We therefore use their original MFCC features in our experiments.
For consistency with our own model, instead of measuring $L_2$ distance on PCA features, as they do, we add an extra hidden layer and use $L_1$ distance.
\vspace{-3mm}
\mypar{Ours, no GAN} Finally, as an ablation, we compare our full model to the prediction of the translation architecture alone, without the adversarial discriminator.

\subsubsection{Evaluation Metrics}
Our main quantitative evaluation metric is the $L_1$ regression loss of the different models in comparison. We additionally report results according to the percent of correct keypoints (PCK)~\cite{pck}, a widely accepted metric for pose detection. Here, a predicted keypoint is defined as correct if it falls within $\alpha \max(h,w)$ pixels of the ground truth keypoint, where $h$ and $w$ are the height and width of the person bounding box, respectively.

We note that PCK was designed for localizing object parts, whereas we use it here for a cross-modal prediction task (predicting pose from audio). First, unlike $L_1$, PCK is not linear and correctness scores fall to zero outside a hard threshold. Since our goal is not to predict the ground truth motion but rather to use it as a training signal, $L_1$ is more suited to measuring how we perform on average. Second, PCK is sensitive to large gesture motion as the correctness radius depends on the width of the span of the speaker's arms. While~\cite{pck} suggest $\alpha=0.1$ for data with full people and $\alpha=0.2$ for data where only half the person is visible, we take an average over $\alpha=0.1,0.2$ and show the full results in the supplementary.

\subsection{Quantitative Evaluation}
We compare the results of our method to the baselines using our quantitative metrics. To assess whether our results are perceptually convincing, we conduct a user study. Finally, we ask whether the gestures we predict are person-specific and whether the input speech is indeed a better predictor of motion than the initial pose of the gesture.

\subsubsection{Numerical Comparison} We compare to all baselines on $2{,}048$ randomly chosen test set intervals per speaker and display the results in Table~\ref{tab-L1}. We see that on most speakers, our model outperforms all others, where our no-GAN condition is slightly better than the GAN one. This is expected, as the adversarial discriminator pushes the generator to snap to a single mode of the data, which is often further away from the actual ground truth than the mean predicted by optimizing $L_1$ loss alone.
Our model outperforms the RNN-based model on most speakers. Qualitatively, we find that this baseline predicts relatively small motions on our data, which may be due to the fact that it has relatively low capacity compared to our \unet{} model.

\begin{table*}
\small
\begin{center}
\setlength{\tabcolsep}{3pt}
\begin{tabular}{lcccccccccccc}
\toprule
Model & Meyers & Oliver & Conan & Stewart & Ellen & Kagan & Kubinec & Covach & Angelica & Almaram & \textbf{Avg. L1} & \textbf{Avg. PCK}  \tabularnewline
\midrule
Median                     & $0.66$ & $0.69$ & $0.79$ & $0.63$ & $0.75$ & $0.80$ & $0.80$ & $0.70$ & $0.74$ & $0.76$ & $0.73$ & $38.11$ \tabularnewline
Random                     & $0.93$ & $1.00$ & $1.10$ & $0.94$ & $1.07$ & $1.11$ & $1.12$ & $1.00$ & $1.04$ & $1.08$  & $1.04$ & $26.55$ \tabularnewline
NN~\cite{vggish}           & $0.88$ & $0.96$ & $1.05$ & $0.93$ & $1.02$ & $1.11$ & $1.10$ & $0.99$ & $1.01$ & $1.06$ & $1.01$ & $27.92$ \tabularnewline
\midrule
RNN~\cite{shlizermanaudio} & $0.61$ & $0.66$ & $0.76$ & $0.62$ & $\mathbf{0.71}$ & $0.74$ & $0.73$ & $0.72$ & $\mathbf{0.72}$ & $\mathbf{0.75}$  & $0.70$ & $39.69$ \tabularnewline
\midrule
Ours, no GAN               & $\mathbf{0.57}$ & $\mathbf{0.60}$ & $\mathbf{0.63}$ & $\mathbf{0.61}$ & $\mathbf{0.71}$ & $\mathbf{0.72}$ & $\mathbf{0.68}$ & $\mathbf{0.69}$ & $0.75$ & $0.76$   & $\mathbf{0.67}$ & $\mathbf{44.62}$ \tabularnewline
Ours, GAN                  & $0.77$ & $0.63$ & $0.64$ & $0.68$ & $0.81$ & $0.74$ & $0.70$ & $0.72$ & $0.78$ & $0.83$ & $0.73$ & $41.95$ \tabularnewline
\bottomrule
\end{tabular}
\end{center}
\vspace{-4mm}
\caption{Quantitative results for the speech to gesture translation task using $L_1$ loss (lower is better) on the test set. The rightmost column is the average PCK value (higher is better) over all speakers and $\alpha=0.1,0.2$ (See full results in supplementary).}
\label{tab-L1}
\end{table*}

\subsubsection{Human Study}
To gain insight into how synthesized gestures perceptually compare to real motion, we conducted a small-scale real vs. fake perceptual study on Amazon Mechanical Turk. We used two speakers who are always shot from the same camera viewpoint: Oliver, whose gestures are relatively dynamic and Meyers, who is relatively stationary.
We visualized gesture motion using videos of skeletal wire frames. To provide participants with additional context, we included the ground truth mouth and facial keypoints of the speaker in the videos. We show examples of skeletal wire frame videos in our video supplementary material.

Participants watched a series of video pairs. In each pair, one video was produced from a real pose sequence; the other was generated by an algorithm---our model or a baseline. Participants were then asked to identify the video containing the motion that corresponds to the speech sound (we did not verify that they in fact listened to the speech while answering the question). Videos of $4$ seconds or $12$ seconds each of resolution $400\times 226$ (downsampled from $910\times 512$ in order to fit two videos side-by-side on different screen sizes) were shown, and after each pair, participants were given unlimited time to respond. We sampled $100$ input audio intervals at random and predicted from them a $2D$-keypoint motion sequence using each method. Each task consisted of $20$ pairs of videos and was performed by $300$ different participants. Each participant was given a short training set of $10$ video pairs before the start of the task, and was given feedback indicating whether they had correctly identified the ground-truth motion. %

We compared all the gesture-prediction models (Section~\ref{sec:baselines}) and assessed the quality of each method using the rate at which its output fooled the participants. Interestingly, we found that for the dynamic speaker all methods that generate realistic motion fooled humans at similar rates. As shown in Table~\ref{tab-mturk}, our results for this speaker were comparable to real motion sequences, whether selected by an audio-based nearest neighbor approach or randomly. For the stationary speaker who spends most of the time in rest position, real motion was more often selected as there is no prediction error associated with it.
While the nearest neighbor and random motion models are significantly less accurate quantitatively (Table~\ref{tab-L1}), they are perceptually convincing because their components are realistic.

\begin{table}
\footnotesize
\centering
\setlength{\tabcolsep}{3pt}
\begin{tabular}{lcccc}
\toprule
 & \multicolumn{2}{c}{Oliver} & \multicolumn{2}{c}{Meyers} \tabularnewline
\midrule
Model & $4$ seconds & $12$ seconds & $4$ seconds & $12$ seconds \tabularnewline
\midrule
Median & $12.1 \pm 2.8$ &  $6.7 \pm 2.0$ &  $34.0 \pm 4.2$ & $25.8 \pm 3.9$\tabularnewline
\midrule
Random &  $\mathbf{34.2 \pm 4.0}$ & $\mathbf{29.1 \pm 3.7}$ & $\mathbf{40.9 \pm 4.6}$  & $\mathbf{34.3 \pm 4.4}$ \tabularnewline
NN~\cite{vggish} & $\mathbf{36.9 \pm 3.9}$ & $\mathbf{26.4 \pm 3.8}$ & $\mathbf{43.5 \pm 4.5}$  & $\mathbf{33.3 \pm 4.4}$  \tabularnewline
\midrule
RNN~\cite{shlizermanaudio} & $18.2 \pm 3.2$ & $10.0 \pm 2.5$ & $37.5 \pm 4.6$   & $19.4 \pm 3.6$ \tabularnewline
\midrule
Ours, no GAN & $25.0 \pm 3.8$ & $19.8 \pm 3.4$ & $36.1 \pm 4.3$  & $\mathbf{33.1 \pm 4.2}$ \tabularnewline
Ours, GAN & $\mathbf{35.4 \pm 4.0}$ & $\mathbf{27.8 \pm 3.9}$ & $33.2 \pm 4.4$  & $22.0 \pm 4.0$\tabularnewline
\bottomrule
\end{tabular}
\caption{Human study results for the speech to gesture translation task on $4$ and $12$-second video clips of two speakers---one dynamic (Oliver) and one relatively stationary (Meyers). As a metric for comparison, we use the percentage of times participants were fooled by the generated motions and picked them as real over the ground truth motion in a two-alternative forced choice. We found that humans were not sensitive to the alignment of speech and gesture. For the dynamic speaker, gestures with realistic motion---whether randomly selected from another video of the same speaker or generated by our GAN-based model---fooled humans at equal rates (no statistically significant difference between the bolded numbers). Since the stationary speaker is usually at rest position, real unaligned motion sequences look more realistic as they do not suffer from prediction noise like the generated ones.
}
\label{tab-mturk}
\end{table}

\subsubsection{The Predicted Gestures are Person-Specific}
\begin{figure}
\centering
\includegraphics[width=0.8\linewidth]{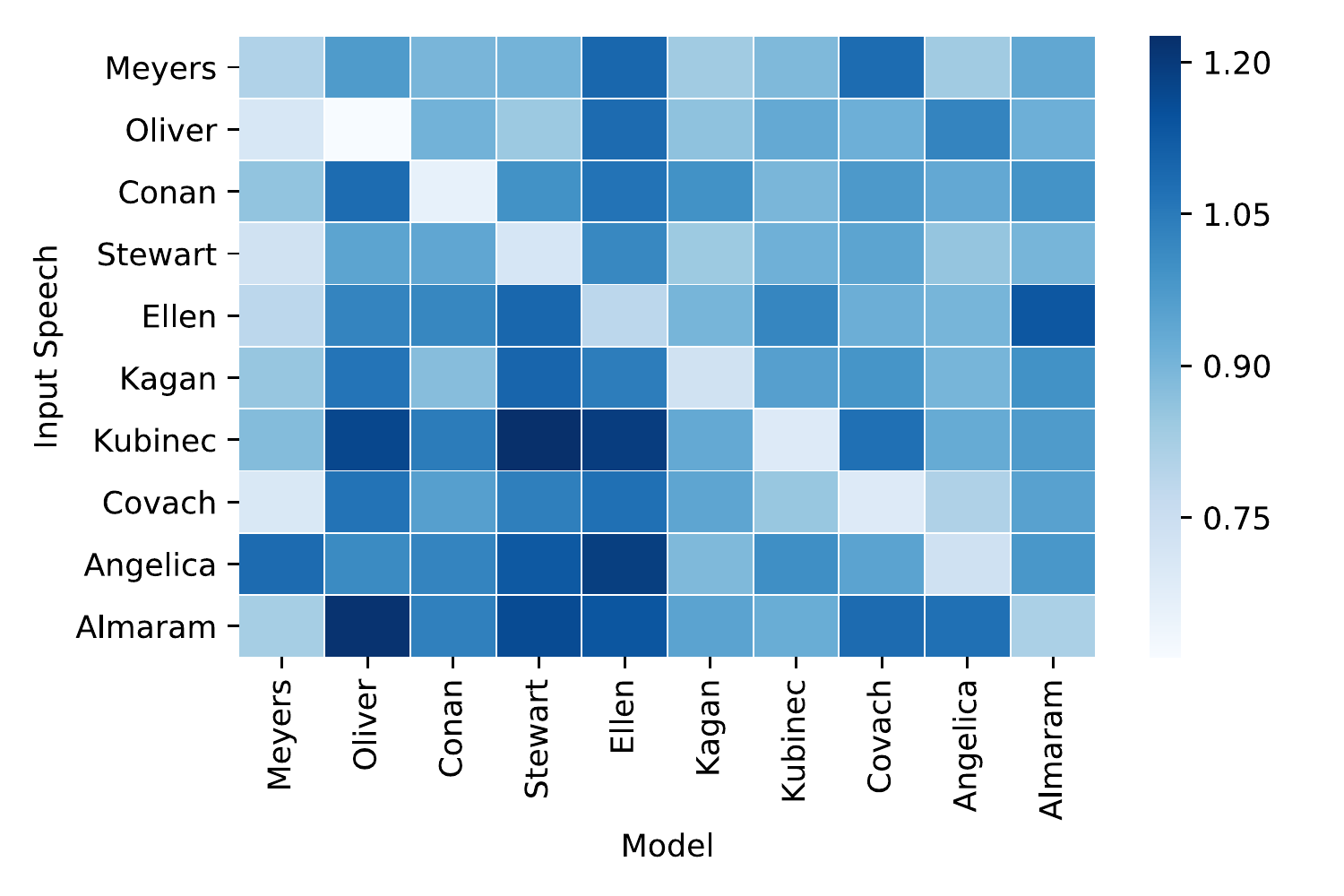}
\vspace{-2mm}
  \caption{Our trained models are person-specific. For every speaker audio input (row) we apply all other individually trained speaker models (columns). Color saturation corresponds to $L_1$ loss values on a held out test set (lower is better). For each row, the entry on the diagonal is lightest as models work best using the input speech of the person they were trained on.}
\label{fig:cross-person}
\end{figure}

For every speaker's speech input (Figure~\ref{fig:cross-person} rows), we predict gestures using all \emph{other} speakers' trained models (Figure~\ref{fig:cross-person} columns). We find that on average, predicting using our model trained on a different speaker performs better numerically than predicting random motion, but significantly worse than always predicting the median pose of the input speaker (and far worse than the predictions from the model trained on the input speaker). The diagonal structure of the confusion matrix in Figure~\ref{fig:cross-person} exemplifies this.

\subsubsection{Speech is a Good Predictor for Gesture}
Seeing the success of our translation model, we ask how much does the audio signal help {\em when the initial pose of the gesture sequence is known}. In other words, how much can sound tell us beyond what can be predicted from motion dynamics. To study this, we augment our model by providing it the pose of the speaker directly preceding their speech, which we incorporate into the bottleneck of the \unet{} (\fig{fig:translation-arch}). We consider the following conditions:~\textit{Predict median pose}, as in the baselines above. \textit{Predict the input initial pose}, a model that simply repeats the input initial ground-truth pose as its prediction. \textit{Speech input}, our model. \textit{Initial pose input}, a variation of our model in which the audio input is ablated and the network predicts the future pose from only an initial ground-truth pose input, and \textit{Speech \& initial pose input}, where we condition the prediction on both the speech and the initial pose.

Table~\ref{tab-correlation} displays the results of the comparison for our model trained without the adversarial discriminator (no GAN). When comparing the \textit{Initial pose input}  and \textit{Speech \& initial pose input} conditions, we find that the addition of speech significantly improves accuracy when we average the loss across all speakers ($p<10^{-3}$ using a two sided t-test). Interestingly, we find that most of the gains come from a small number of speakers (\eg Oliver) who make large motions during speech.

\begin{table}
\small
\centering
\setlength{\tabcolsep}{3pt}
\begin{tabular}{clcc}
\toprule
\multicolumn{2}{c}{Model} & Avg. $L_1$ & Avg. PCK  \tabularnewline
\midrule
\parbox[t]{4mm}{\multirow{2}{*}{\rotatebox[origin=l]{90}{Pred.}}}
& Predict the median pose & $0.73$ & $38.11$ \tabularnewline
& Predict the input initial pose  & $0.53$ & $60.50$\tabularnewline
\midrule
\parbox[t]{4mm}{\multirow{3}{*}{\rotatebox[origin=c]{90}{Input}}}
& Speech input & $0.67$  & $44.62$ \tabularnewline
& Initial pose input  & $0.49$ & $61.24$ \tabularnewline
& Speech \& initial pose input & $\mathbf{0.47}$ & $\mathbf{62.39}$ \tabularnewline
\bottomrule
\end{tabular}
\caption{How much information does sound provide once we know the initial pose of the speaker? We see that the initial pose of the gesture sequence is a good predictor for the rest of the 4-second motion sequence (second to last row), but that adding audio improves the prediction (last row). We use both average $L_1$ loss (lower is better) and average PCK over all speakers and $\alpha=0.1,0.2$ (higher is better) as metrics of comparison. We compare two baselines and three conditions of inputs.}
\label{tab-correlation}
\end{table}

\subsection{Qualitative Results}
\begin{figure*}
\centering
\includegraphics[width=\linewidth]{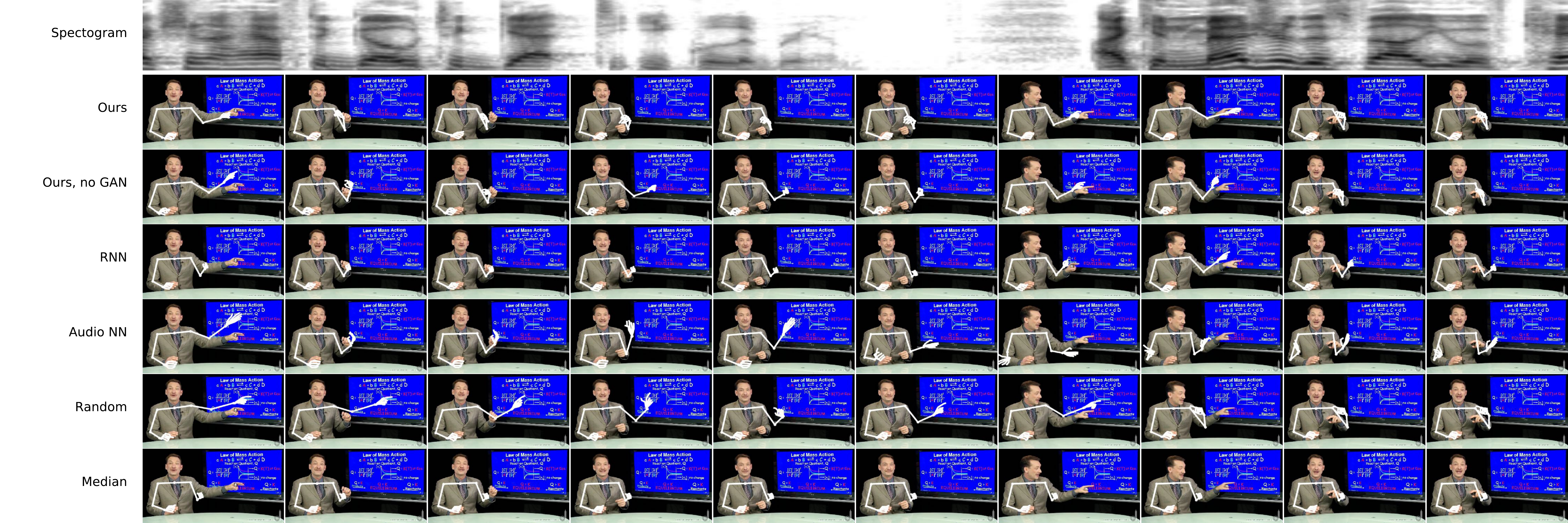} \\
\vspace{3mm}
\includegraphics[width=\linewidth]{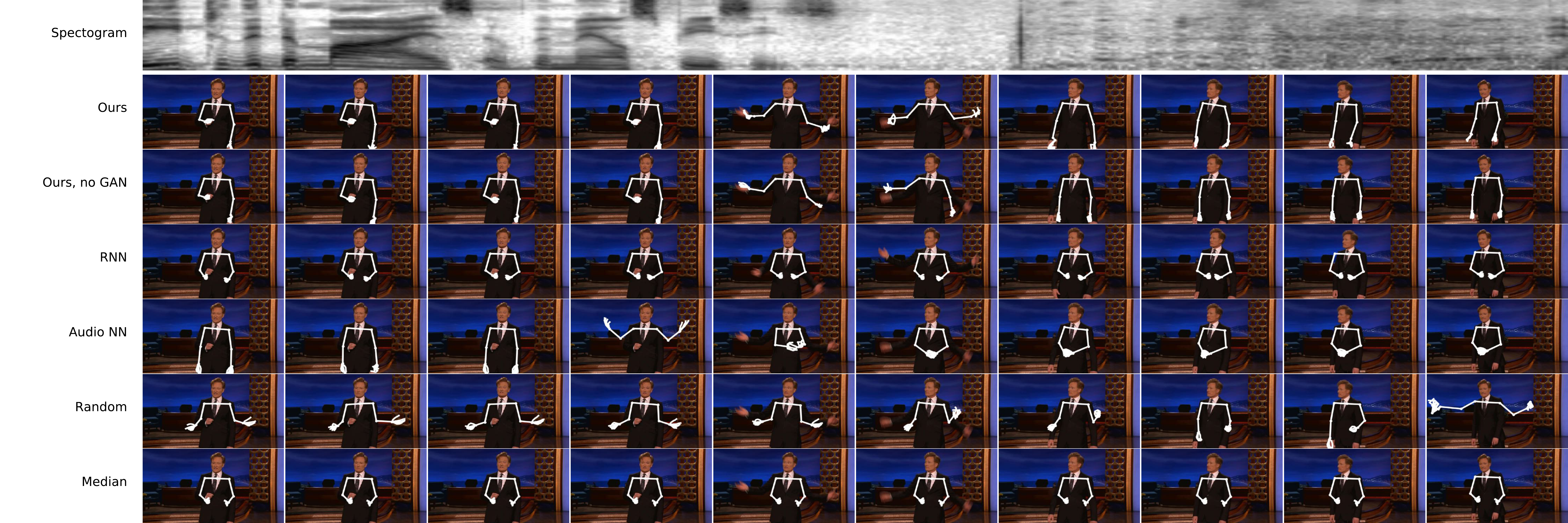} \\
  \caption{{\em Speech to gesture translation qualitative results.} We show the input audio spectrogram and the predicted poses overlaid on the ground-truth video for Dr. Kubinec (lecturer) and Conan O'Brien (show host). {\em See our supplementary material for more results.}}
\label{fig:translation-qualitative}
\end{figure*}

We qualitatively compare our speech to gesture translation results to the baselines and the ground truth gesture sequences in Figure~\ref{fig:translation-qualitative}. Please refer to our supplementary video results which better convey temporal information.

\section{Conclusion}

Humans communicate through both sight and sound, yet the connection between these modalities remains unclear~\cite{kendon_2004}. In this paper, we proposed the task of predicting person-specific gestures from ``in-the-wild" speech as a computational means of studying the connections between these communication channels. We created a large person-specific video dataset and used it to train a model for predicting gestures from speech. Our model outperforms other methods in an experimental evaluation.

Despite its strong performance on these tasks, our model has limitations that can be addressed by incorporating insights from other work. For instance, using audio as input has its benefits compared to using textual transcriptions as audio is a rich representation that contains information about prosody, intonation, rhythm, tone and more. However, audio does not directly encode high-level language semantics that may allow us to predict certain types of gesture (\eg metaphorics), nor does it separate the speaker's speech from other sounds (\eg audience laughter). Additionally, we treat pose estimations as though they were ground truth, which introduces significant amount of noise---particularly on the speakers' fingers.

We see our work as a step toward a computational analysis of conversational gesture, and opening three possible directions for further research. The first is in using gestures as a representation for video analysis: co-speech hand and arm motion make a natural target for video prediction tasks. The second is using in-the-wild gestures as a way of training conversational agents: we presented one way of visualizing gesture predictions, based on GANs~\cite{Chan2018dance}, but, following classic work~\cite{cassell2000embodied}, these predictions could also be used to drive the motions of virtual agents. Finally, our method is one of only a handful of initial attempts to predict motion from audio. This cross-modal translation task is fertile ground for further research. 

\vspace{3mm}
\noindent
{\small{\bf Acknowledgements:} This work was supported, in part, by the AWS Cloud Credits for Research and the DARPA MediFor programs, and the UC Berkeley Center for Long-Term Cybersecurity. Special thanks to Alyosha Efros, the bestest advisor, and to Tinghui Zhou for his dreams of late-night talk show stardom.}

{\small
\bibliographystyle{ieee}
\bibliography{main}
}

\section{Appendix}
\subsection{Dataset}
\paragraph{Data collection and processing}  We collected internet videos by querying YouTube for each speaker, and de-duplicated the data using the approach of~\cite{fouhey2017lifestyle}. We then used out-of-the-box face recognition and pose detection systems to split each videos into intervals in which only the subject appears in frame and all detected keypoints are visible. Our dataset consists of $60{,}000$ such intervals with an average length of $8.7$ seconds and a standard deviation of $11.3$ seconds. In total, there are $144$ hours of video. We split the data into $80\%$ train, $10\%$ validation, and $10\%$ test sets, such that each source video only appears in one set.

\mypar{Quality of dataset annotations} We estimate whether the accuracy of the pseudo ground truth is good enough to support our quantitative conclusions via the following experiment.
We took a $200$-frame subset of the pseudo ground truth used for training and had it labeled by $3$ human observers with neck and arm keypoints. We quantified the consensus between annotators via, $\sigma_i$, a standard deviation per keypoint-type $i$, as is typical in COCO~\cite{coco} evaluation. We also computed $||op_i - \mu_i||$, the distance between the OpenPose detection and the mean of the annotations, and $||prediction - \mu_i||$ the distance between our audio-to-motion prediction and the annotation mean. We found that \emph{the pseudo ground truth is close to human labels}, since $0.14 = E[||op_i - \mu_i||] \approx E[\sigma_i] = 0.06$; And that \emph{the error in the pseudo ground truth is small enough for our task}, since $ 0.25 = ||prediction - \mu_i|| >> \sigma_i = 0.06$. Note that this is a \emph{lower bound} on the prediction error since it is computed on training data samples.

\subsection{Learning Individual Gesture Dictionaries}
\begin{figure}
  \centering \includegraphics[width=\linewidth]{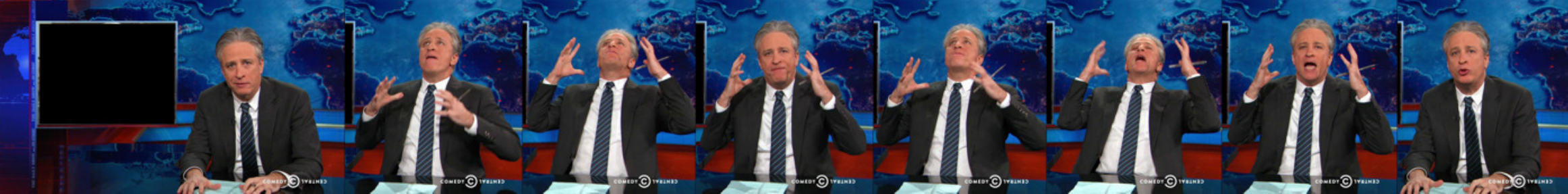}
  \caption{A segmented gesture unit.}
  \label{fig:example-gesture}
\end{figure}

\begin{figure}
\begin{center}
\includegraphics[width=\linewidth]{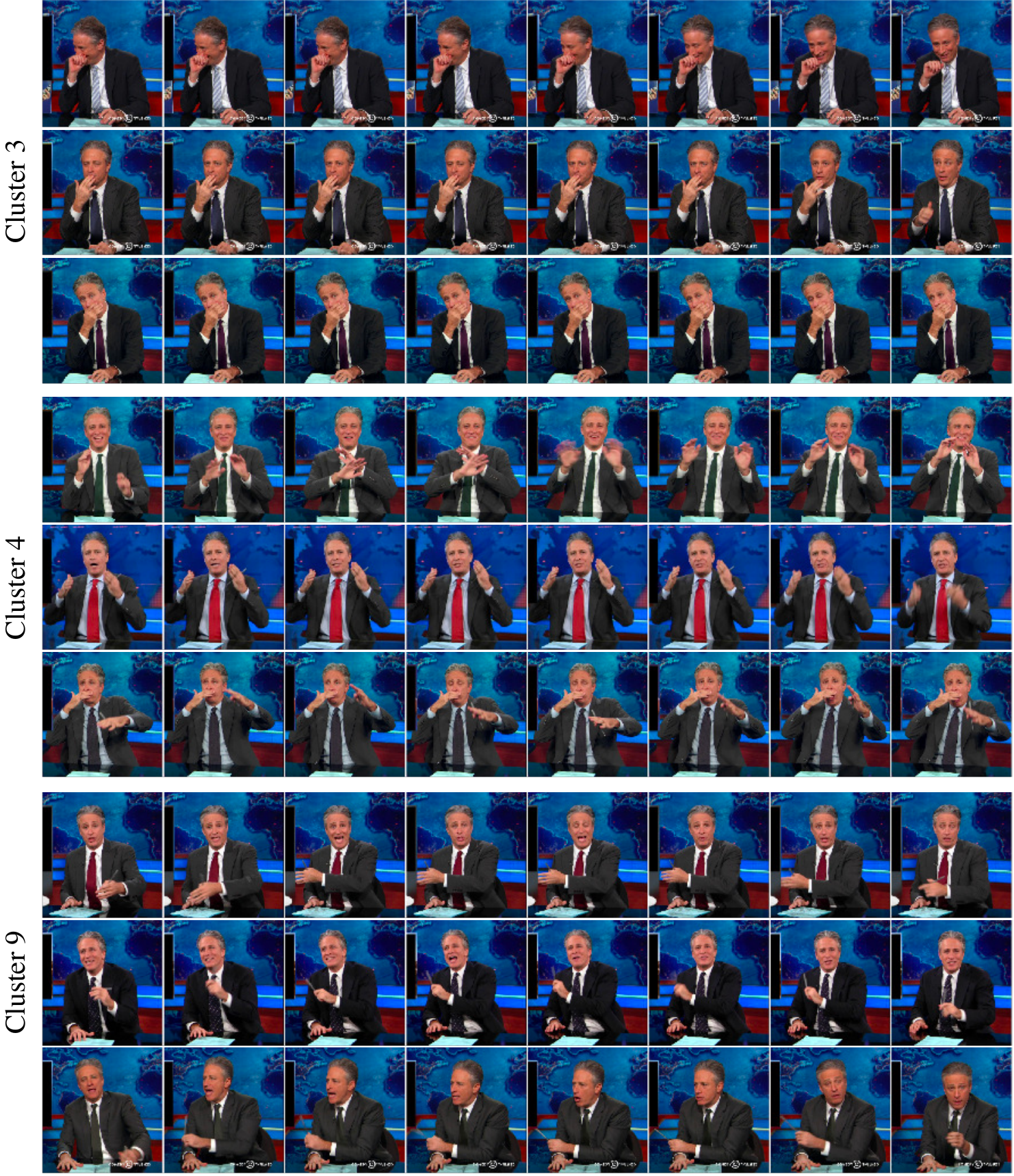}\end{center}
  \caption{\emph{Individual styles of gesture}. Examples from Jon Stewart's gesture dictionary.}
\label{fig:space-of-gesures}
\end{figure}

\paragraph{Gesture unit segmentation}
We use an unsupervised method for building a dictionary of an individual's gestures. We segment motion sequences into gesture units, propose an appropriate descriptor and similarity metric and then cluster the gestures of an individual.

A \textit{gesture unit} is a sequence of gestures that starts from a rest position and returns to a rest position only after the last gesture~\cite{kendon_2004}. While~\cite{McNeill92} observed that most of their subjects usually perform one gesture at a time, a study of an $18$-minute video dataset of TV speakers reported that their gestures were often strung together in a sequence~\cite{Kipp-synthesis}. We treat each gesture unit -- from rest position to rest position -- as an atomic segment.

We use an unsupervised approach to the temporal segmentation of gesture units based on prediction error (by contrast,~\cite{SVN-segmentation} use a supervised approach). Given a motion sequence of keypoints (Section~\ref{sec:data}) from time $t_0$ to $t_T$, we try to predict the $t_{T+1}$ pose. A low prediction error may signal that the speaker is at rest, or that they are in the middle of a gesture that the model has frequently seen during training. Since speakers spend most of the time in rest position~\cite{kendon_2004}, a high prediction error may indicate that a new gesture has begun. We segment gesture units at points of high prediction error (without defining a rest position per person). An example of a segmented gesture unit is displayed in Figure~\ref{fig:example-gesture}.  We train a segmentation model per subject and do not expect it to generalize across speakers.

\mypar{Dictionary learning} We use the first $5$ principal components of the keypoints computed over all static frames as a gesture unit descriptor. This reduces the dimensionality while capturing $93\%$ of the variance. We use dynamic time warping~\cite{dtw} as our distance metric to account for temporal variations in the execution of similar gestures. Since this is not a Euclidean norm, we must compute the distance between each pair of datapoints. We precompute a distance matrix for a randomly chosen sample of $1,000$ training gesture units and use it to hierarchically cluster the datapoints. 

\mypar{Individual styles of gesture} These clusters represent an unsupervised definition of the typical gestures that an individual performs. For each dictionary element cluster we define the central point as the point that is closest on average to all datapoints in the cluster. We sort the gesture units in each cluster by their distance to the central point and pick the most central ones for display. We visualize some examples of the dictionary of gestures we learn for Jon Stewart in Figure~\ref{fig:space-of-gesures}. 

\end{document}